\documentclass[10pt,conference,a4paper]{IEEEtran}
\IEEEoverridecommandlockouts
\usepackage{graphicx,booktabs,amsmath,amssymb,url,color,flushend}
\usepackage[linesnumbered,ruled]{algorithm2e}
\usepackage[misc]{ifsym}

\usepackage{subfig}
\usepackage{floatrow}

\usepackage{bm}
\usepackage{multirow}
\usepackage[noend]{algpseudocode}
\def\BibTeX{\rm B\kern-.05em{\sc i\kern-.025em b}\kern-.08em
		T\kern-.1667em\lower.7ex\hbox{E}\kern-.125emX}
\begin{document}
\title{Classification of Long Noncoding RNA Elements Using Deep Convolutional Neural Networks and Siamese Networks}

\author{\IEEEauthorblockN{Brian McClannahan$^\dag$, Cuncong Zhong$^\dag$, Guanghui Wang$^\ddag$}
	\IEEEauthorblockA{\textit{$^\dag$ Department of Electrical Engineering and Computer Science, The University of Kansas,
			Lawrence, KS, USA 66045}} 
	\IEEEauthorblockA{\textit{$^\ddag$ Department of Computer Science, Ryerson University, Toronto, ON, Canada M5B 2K3}}}

\maketitle
\begin{abstract}
In the last decade, the discovery of noncoding RNA (ncRNA) has exploded. Classifying these ncRNA is critical to determining their function. This thesis proposes a new method employing deep convolutional neural networks (CNNs) to classify ncRNA sequences. To this end, this paper first proposes an efficient approach to convert the RNA sequences into images characterizing their base-pairing probability. As a result, classifying RNA sequences is converted to an image classification problem that can be efficiently solved by available CNN-based classification models. This research also considers the folding potential of the ncRNAs in addition to their primary sequence. Based on the proposed approach, a benchmark image classification dataset is generated from the RFAM database of ncRNA sequences. In addition, three classical CNN models and three Siamese network models have been implemented and compared to demonstrate the superior performance and efficiency of the proposed approach. Extensive experimental results show the great potential of using deep learning approaches for RNA classification.
\end{abstract}

%\\\\

\section{Introduction}

Ribonucleic Acid (RNA) plays vastly many different roles within living cells. In 2007, with the development of high-throughput sequencing, a large amount of new RNA have been discovered \cite{graphclust}. Many of these RNA were found not to be involved in protein creation and are thus called noncoding RNA (ncRNA). However, just because these RNA do not directly code for protein does not reduce their impact in the cell \cite{review}. Noncoding RNA have been shown to play many important roles in cell processes such as to effect high-order chromosomal dynamics, telomere biology, subcellular structural organization, and gene expression. These processes are central to life \cite{hombach2016non}. Not just in animals, but in plants, fungi, and prokaryotes.

Some significant ncRNA include microRNA (miRNA), which performs posttranscriptional regulation of gene expression. Micro-RNA do this by partial complimentary base pairing to specific messenger RNA (mRNA). This process inhibits protein translation and can also facilitate degradation of the target mRNA \cite{mercer2009long}. Small nuclear RNA (snRNA) also play a pivotal role in gene expression. Small nuclear RNA are involved in the pretranscriptional splicing process. Transfer RNA, another ncRNA, decore the mRNA sequence into peptides or proteins. Ribsomal RNA (rRNA) form the framework of ribosomes, a macromolecular structure essential for protein translation. 

Small non-coding RNA have been proven to affect many different processes in animal development. Small ncRNA function as developmental switches by silencing unwanted messages resulting from leaky transcription or previous synthesis. This acts as maintenance on target mRNA expression, keeping them in optimal ranges \cite{stefani2008small}. Micro RNA play a role in early embryonic development by modulating nodal levels in early embryos. Nodal is a transforming growth factor protein that plays a very important role in early embryonic development. Micro RNA also play key roles in muscle development by regulating proliferation to differentiation switching in muscles and playing an important role in cardiac function \cite{stefani2008small}. It is also important in lymphocyte development as with miR-181, an RNA that increases the sensitivity of DP cells to stimulation of the T-Cell receptor, a type of white blood cell \cite{stefani2008small}. 

Long non-coding RNA (lncRNA) are a subset of ncRNA which are longer than 200 nucleotides. These RNA show very low sequence conservation \cite{ma2013classification}. These lncRNA are the most functionally diverse class of RNA. They are predominantly localized in the nucleus of the cell and specifically within chromatin, the material that forms chromosomes within the nucleus \cite{ma2013classification}. The cascading transcription of lncRNA across the fructose bisphosphate promoter in yeast is associated with the progressive opening of chromatin, increasing access to transcriptional activators \cite{ma2013classification}. Long ncRNA can regulate epigenetic modification and gene expression \cite{review}. Other lncRNA have also been shown to affect post-transcriptional regulation through splicing regulation and transitional control \cite{ma2013classification}. These families of RNA are medically significant because of their ability to regulate associated protein-coding genes. This regulation could contribute to disease if misexpressed, thereby deregulating genes of clinical significance \cite{mercer2009long}. The classification of ncRNAs seeds to categorize ncRNA elements into families based on their sequence and structure to facilitate their functional annotation and prediction \cite{review}.

RNA are defined by three graphical representations: primary structure, secondary structure, and tertiary structure. The primary structure is defined by its sequence. The secondary structure is defined by the bonds formed from folding. The tertiary structure is defined by the shape the RNA takes in the physical world. Given an RNA's sequence, it is much simpler to determine its secondary structure than tertiary structure \cite{herschlag1995rna}. Given this, bioinformatics focuses on the primary and secondary structure. The molecular function of ncRNA is implemented through both of its sequence and structure \cite{graphclust}. Classification of ncRNAs purely using their sequences is insufficient, as ncRNAs with conserved secondary structures may share low sequence identity due to the presence of co-variant mutations. Many ncRNA cannot be aligned by purely Sequence-based alignment techniques struggle when the pairwise sequence identities are less than 60\%. Initial methods looked at using structure-based alignment techniques like MARNA \cite{marna} and RNAforester \cite{rnaforester}, but these struggled because of poor accuracy of single structure predictions \cite{locarna}. Therefore, modern methods for ncRNA clustering use both the primary sequence and secondary structure features \cite{dotaligner}.

Unfortunately, the consideration of the secondary structure increases the time complexity for pairwise ncRNA comparison, from $O(l^2)$ with pure sequence to $O(l^4)$ with both sequence and secondary structure. The high time complexity thus makes clustering of a large amount of ncRNA elements infeasible with current methods. New methods are needed to handle the continuing growth of discovered ncRNA. The initial release of RFAM 1.0 in 2003 contained 25 unique RNA families. In 2018, RFAM 13.0 was released, containing 2,687 families. Faster algorithms are needed to process the exponentially growing size of bioinformatics data, especially within ncRNA.

Given a newly-discovered RNA, modern day approaches rely on biologists to manually align sequences with other RNA across similar-length RNA families in order to determine their RNA family. This is a very time-consuming process. The obvious solution is to use sequence alignment approaches to determine RNA families. These sequence alignment techniques use dynamic programming which have high complexity, especially on longer RNA. It also requires the RNA in the same family to have similar sequence identities, of which many lncRNA do not. Calculating the RNA secondary structure while aligning the sequences improves the classification accuracy of lncRNA but increases the complexity of the classification from $O(l^2)$ to $O(l^4)$. The other option is to use structural alignment, but these methods have previously shown poor accuracy. Thus, it is necessary to find quicker algorithms that can accurately classify RNA with just its secondary structure, and not only determine if an RNA is within a currently known family, but can determine if the RNA should be defined within a completely new RNA family.

In this paper, we propose a new approach to general RNA family classification using image processing techniques. For a given ncRNA element, the base-pairing probability matrix (BPPM) was computed using RNAfold from the Vienna Package. We convert the BPPM into a gray-scale image, using the intensity of each pixel to represent the base-pairing probability of the corresponding bases. We then apply three different deep CNN algorithms (VGGnet-19, ResNet-50, and ResNet-101) to classify these images. These deep CNN algorithms are then wrapped into Siamese networks for classification as well. We tested this approach using the RFAM database \cite{rfam}, and showed an 85\% classification accuracy with the CNN model and 79\% classification accuracy with the Siamese model.

In this paper, we propose:
\begin{enumerate}
    \item To convert the problem of RNA sequence classification into an image classification problem.
    \item An efficient approach to convert two RNA sequences into an image and generate an image dataset for RNA sequences from the same and different families.  
    \item Implement three classical deep learning-based classification models and compare their performance in RNA classification. The results demonstrate the feasibility of the advantages of the proposed approach.
    \item Implement these three classical deep learning-based classification models as Siamese Networks to evaluate their performance with the deep CNNs as feature-extractors on ncRNA.
\end{enumerate}

A part of this paper has been published at the 2020 IEEE International Conference on Systems, Man, and Cybernetics \cite{mcclannahan2020classification}.

\section{Related Works}

\textbf{Analyzing RNA by Secondary Structure} Initial methods, like MARNA and RNAforester, utilized just the predicted or actual secondary structure for classifying ncRNA \cite{locarna}. MARNA used pairwise-sequence alignments to make their RNA structural predictions. These structural predictions were then used in multi-structural alignments \cite{marna}. RNAforester built an RNA secondary structure alignment algorithm by representing RNA secondary structures as trees and aligning them with tree alignment algorithms. However, these methods were found to struggle with accurate predictions. There are several other, more modern methods that are used for classifying RNA by its secondary structure. QRNA utilizes a context-free grammar constrained by structural RNA evolution paired with two different hidden Markov models for determining other constraints\cite{qrna}. RNAfold \cite{rnafold} and Mfold \cite{mfold} use secondary structure comparisons based on a minimum free energy model. RNAz predicts the conserved structure of RNA based on its thermodynamic stability \cite{rnaz}. Sankoff-style folding algorithms have also shown effectiveness in doing simultaneous primary and secondary structure alignments \cite{sankoff}. CARNA showed how these could be used to classify RNA by finding a predicted secondary structure during RNA alignment \cite{carna}. Several other approaches compare multiple graph representations of RNA secondary structure such as GraphClust \cite{graphclust} and \cite{graph_2004}.

The most successful methods have combined sequence alignment techniques with a partition function, like the McCaskill Partition function \cite{mccaskill}, to compare the primary and secondary structures at the same time, such as LocaRNA-P \cite{locarna}. LocaRNA uses the McCaskill partition function initially to generate a base-pair probability matrices for two given RNA. These matrices are then scored based on confidence in the individual alignment columns and the predicted consensus structure, called STARs (sequence-structure-based alignment reliability). These STARs are calculated by dividing the Boltzmann weights by the total partition function. STARs are then used to do multiple alignments across RNA to generate STAR profiles. These profiles predict the boundaries of local regions of conserved sequence and structure to find ncRNA in alignments of longer sequences. These star profiles allow for RNA clusters to be found when pointed at a genome \cite{locarna}.

LocaRNA was one of the first classifiers not to struggle to accurately analyze ncRNA. Another more recent approach built with LocaRNA's approach in mind is DotAligner, which also uses both the primary sequence and secondary structure to discover sequence motifs, but is more focused on lncRNAs \cite{dotaligner}. DotAligner first generates a dynamic programming matrix filled by aligning the two given sequences based on their similarity and cumulative base-pair probabilities. Then the structural compatibility of the two RNA sequences is calculated using a partition function over all pairwise alignments. The two most likely structural ensembles generated by the partition function are considered. Lastly, the sequence alignments are warped towards the alignment that includes structural features, thereby exposing the common structural features hidden by suboptimal base-pairing ensembles.

\textbf{Image Classification with Convolutional Neural Networks} Recently, researchers have applied CNNs to many different fields of applications, including image classification \cite{cen2019boosting}, translation \cite{xu2019adversarially}, object detection \cite{mo2018efficient}\cite{ma2020mdfn}, depth estimation \cite{he2018spindle}, crowd counting \cite{sajid2020zoomcount}, and medical image analysis \cite{patel2020comparative}. AlexNet was the first CNN to be applied towards image classification tasks. AlexNet used a series of overlapping convolutional masks of size 11x11, 5x5, and 3x3 pixels over the image. Convolutional layers were followed by pooling layers and ReLU activation layers. ReLU is an important activation function in Image classification because it adds non-linearity and also removes negative values. Pixel values can never be lower than 0 so ReLU purges any weights that generate negative values while maintaining positive values coming out of the convolution masks. AlexNet won the ImageNet Large Scale Visual Recognition Challenge (ILSVRC) 2012 competition with a top-5 error rate of 15.3\%, significantly outperforming other top performing methods which only achieved 26.2\% top-5 error rate \cite{alexnet}. GoogleNet followed up on AlexNet's innovation by introducing the inception module. The inception module employs $3 \times 3$ and $5 \times 5$ convolution masks followed by $1 \times 1$ convolution masks for dimensionality reduction and reduces the total number of parameters required to be trained by the network. GooglNet achieved a top-5 error rate of 6.67\% at the ILSVRC 2014 classification task \cite{googlenet}. 

VGGnet was released the same year as GoogleNet. VGGnet demonstrated a method of building CNNs with significantly fewer parameters to be trained. By stacking $3 \times 3$ convolutional filters, much larger convolutional filters can be implemented more efficiently. For example, a $7 \times 7$ convolutional filter can be implemented by stacking three $3 \times 3$ convolutional filters on top of each other. For $C$ channels in the convolutional layer, the $7 \times 7$ convolutional filter reduces from $7^2C^2$ = $49C^2$ parameters to $3(3^2C^2)$ = $27C^2$ parameters, a 45\% reduction in parameters with the same output. VGGnet achieved the first place in localization and the second in classification at ImageNet Challenge 2014 with a top-5 error rate of 7.3\% \cite{vggnet}. ResNet introduced residual mappings to CNNs. A residual mapping within a neural network layer approximating the function $H(x)$ changes the output of the layer from $F(x) = H(x)$ to $F(x) = H(x) + x$. When a layer with a residual mapping is trained, if it fails to create an accurate approximation of $F(x)$ the output of the layer is still just $x$ instead of a warped $H(x)$ that corrupts the rest of the network. This change allows significantly deeper networks to be trained without overfitting. The largest VGGnet model is 19 layers and has 19.6 billion floating point operations (FLOPs). ResNet was able to build significantly deeper models including a network with 152 layers. Since the network was able to be much deeper due to residual mappings utilized, the network was able to be much thinner and the 152 layer network only had 11.3 billion FLOPs, 40\% FLOPs compared to the 19-layer VGGnet model. Using these improvements, ResNet achieved first in the ILSVRC 2015 classification task with a top-5 error rate of 6.71\% \cite{resnet}. After ResNet created significantly deeper networks, wide residual networks showed that similar results could be achieved by greatly increasing the number of parameters in a shallow network and utilizing dropout layers \cite{wrn}. Other approaches include DenseNet, a CNN designed to compute dense, multi-scale feature pyramids using up and down-sampling of image patches with a CNN-based object classifier \cite{iandola2014densenet}.

Very few studies have been performed for RNA classification using CNN based classifiers. CNNclust \cite{cnnclust} utilized one-dimensional CNNs for ncRNA sequence motif discovery. Similar to DotAligner, CNNclust used both the sequence and the secondary structure to more accurately find sequence motifs and then use those to make accurate classifications. However, CNNclust utilized one-dimensional convolutions over the primary sequence and base-pair probability as opposed to using partition functions like LocaRNA and DotAligner. This one-dimensional CNN took in a one-hot encoded sequence value of the alignment not the sequence, allowing for gap values to be included in the input. Along with the sequence, a three-dimensional vector was also input to the network defining the secondary structure at that nucleotide. This vector was a one-hot encoded vector representing whether the current nucleotide would bond upstream, downstream, or not at all. The network then performed sets of convolution operations over the sequence and structure representations followed by max pooling layers and then three fully-connected layers to classify two RNA. All classifications are then combined into an adjacency matrix and family clusters are then extracted from the adjacency matrix such that each family is defined by the complete maximal sub-graphs extracted.

\textbf{Siamese Networks} Siamese neural networks (SiNN) are more complicated neural networks that try to take advantage of other network architectures to make predictions about similarity. A Siamese network has two branches with each branch sharing the exact same architecture and weights. The network receives two inputs and feeds each input through one of the branches, but the exact order of the input is irrelevant to the final output of the SiNN. Each output from the branch networks are then passed through a comparison layer that evaluates the output of each for similarity. This similarity is then passed to a top network which makes the final classification. Siamese Neural Networks make use of this twin branch network structure to realize a non-linear embedding from its input domain to some euclidean domain \cite{roy2019siamese}. Effectively the branch network operates as a descriptor computation module and the top network as a input similarity function \cite{zagoruyko_komodakis_2015}. These properties of feature extraction and similarity comparisons make Siamese Networks great at classifying similarity between inputs. 

In \cite{melekhov_kannala_rahtu_2016}, Siamese networks have demonstrated measured improvements over CNNs in image matching. Siamese networks built on top of CNNs have demonstrated the ability to match landmarks more accurately than using a standard CNN, even when the landmarks have not been seen before and the training set contained mislabeled data \cite{melekhov_kannala_rahtu_2016}. Siamese Networks have also been shown to be superior to CNNs in general image matching \cite{zagoruyko_komodakis_2015}. Building off of their improved image matching capabilities, Researchers have also applied Siamese networks to object tracking. It has been demonstrated that Siamese networks use available data more efficiently and perform more efficient spatial searches compared to more generic CNN methods \cite{bertinetto2016fully}. Other approaches have also utilized Siamese networks in person tracking \cite{he2018twofold}. They have increased performance in identifying people across multiple images by identifying features of tracked people and matching them to people in subsequent photos. In fact, Siamese Networks using CNNs were able to show state-of-the-art performance in generic image matching with only a single initial observation \cite{tao_gavves_smeulders_2016}. Siamese networks have shown capabilities outside of image comparisons. In \cite{koch2015siamese}, Siamese networks were applied to one-shot image classification and demonstrated the capability to do one-shot classification for verification. These capabilities extend to other domains too, specifically image classification \cite{koch2015siamese}. Within bioinformatics, Siamese networks have seen use with chromosomal classification to improve performance compared to deep learning models \cite{jindal2017siamese}.

\section{Dataset}

\begin{figure}[htp]
    \centering
	\subfloat{\includegraphics[width=10em]{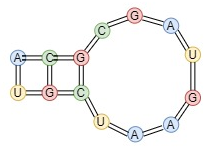}}\\[-8pt]
	\subfloat{\includegraphics[width=10em]{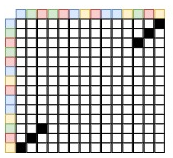}}
	\caption{Top: an example of interior loop of an RNA sequence. Bottom: Corresponding dot-plot matrix of the sequence.}
\end{figure}

\textbf{CNN Dataset}
The dot-plot matrix is generated by letting any cell $(i,j)$ represent the probability that a bond exists between the $i$th and $j$th nucleic acid in the RNA secondary structure (excluding bonds amongst neighboring nucleic acids). Generating a dot-plot matrix from the secondary structure defined by the BPPM gives a matrix that is symmetric along the diagonal, as shown in Fig. 2. The dot-plot matrix is then converted into an image by treating each cell as a pixel where the probability of a bond forming is treated as an intensity between 0 and 1, as illustrated in Fig. 1. This process creates a grey-scale, symmetric image representation of the secondary structure of any given RNA. 

\begin{figure}[htp]
    \centering
	\captionsetup[subfigure]{labelformat=empty}
	\subfloat[a]{\includegraphics[width=10em]{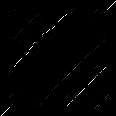}}
	\qquad
	\subfloat[b]{\includegraphics[width=10em]{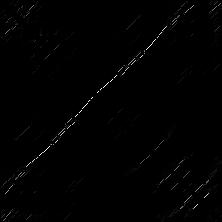}}\\
	\subfloat[c]{\includegraphics[width=10em]{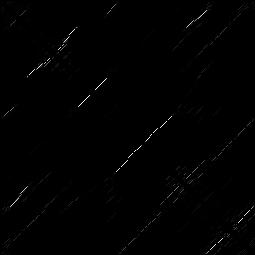}}
	\qquad
	\subfloat[d]{\includegraphics[width=10em]{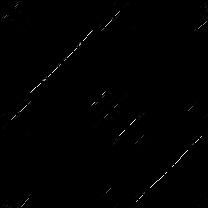}}
	\caption{Four dot-plot examples of RNA secondary structure  RFAM families a) 5S rRNA, b) C0719, c) snR63, d) S\_pombe\_snR42}
	\label{ref_label_overall}
\end{figure}

There are two options for designing the dataset for RNA classification. The first option is to evaluate the dot-plot matrix and classify each RNA with its familial class. Most CNNs are trained and tested on the ImageNet dataset to evaluate their capability in image classification. ImageNet uses 1000 different classes that the CNNs classify on. With this design, the convolution layers detect features and then the fully-connected layers evaluate these features to determine which of the predetermined classes the image is a part of. This would limit the network to only classifying RNA families it has previously been trained on and thus prevent previously unknown RNA families from being defined. The other option is to let each classification made by the network determine if two RNA sequences are from the ``same family" or ``different family".  To solve this problem without altering the RNA dot-plot matrices in some way, it would be necessary to build a network large enough to accept two full images for the two RNA sequences. This would require significant augmentation of current CNN frameworks and would likely fail to take advantage of the powerful feature extraction of CNNs. Using a standard CNN in this capacity would also leave the network irregular (not-square), which, along with being hard to implement and train the network, would take a very long time to train for RNA sequences, and would result in features not being weighed equally due to the CNN no longer being symmetrical. Instead, in this study, we propose a different method to convert the RNA classification problem to an image classification problem. As illustrated in Fig. 2, we make use of the symmetric property of the dot-plot of each sequence and generate one single image from two RNA sequences The bottom-left half from sequence 1 and the top-right half from sequence 2 are stitched together. In this way, the problem is converted from a two-input problem to a single input, ideal for CNNs. Using this method of stitching two dot-plot matrices together, we can make use of all available image classification models from RNA classification. As seen in Fig.3, the images from the same family have better symmetric property than those from different families, however, the differences are very small. The models are trained to distinguish these small changes. Some generated examples are shown in Figures 3.4 and 3.5.

\begin{figure}[ht]
    \centering
	\captionsetup[subfigure]{labelformat=empty}
	\subfloat[]{\includegraphics[width=17em]{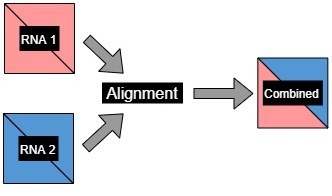}}
	\caption{Combine two dot-plot matrices into one image for classification.}
\end{figure}

To train and evaluate each deep learning model, a large dataset is required. The RFAM database has over 3,000 families. For these experiments, only RNA families that have a sequence length between 200 and 260 were included for the convenience of study. RNA families typically have similar sequence lengths so using a small difference in sequence length is better. It allows the results to accurately reflect how this method would be used in a functional scenario. The RNA dot-plots for all sequences were resized to $224 \times 224$ to combine each pair of them into an image. Using a sequence length between 200 and 260 also adds the benefit that the images do not get warped during resizing, allowing better evaluation of the effectiveness of convolutions in RNA classification. There were 168 families in this range. The 168 families were split into the train/val/test set at a ratio of 70:10:20, resulting in a family split of 121/19/28, as shown in Table 3.1. If the dataset was created exhaustively to cover all possible RNA combinations, the resulting dataset had over 17 million images. A dataset of that size was not necessary for model training as there are a lot of redundant information in the dataset. The largest family size in this set was 712 while many of the families only had 2. Instead of using every possible combination, each family larger than 30 RNA was truncated by randomly selecting 30 RNA from the family. This also has the added benefit of balancing the varying family sizes, preventing the network from favoring classifying RNA with large family sizes as ``Same Family'' more frequently than small RNA families, focusing the model on learning features of RNA. This reduced the total number of images to slightly greater than 2.4 million. 

\begin{figure}[htp]
    \centering
	\captionsetup[subfigure]{labelformat=empty}
	\subfloat[]{\includegraphics[width=10em]{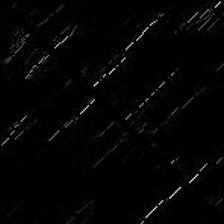}}
	\quad
	\subfloat[]{\includegraphics[width=10em]{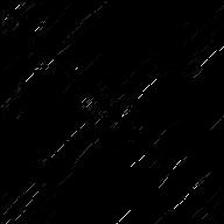}}
	\caption{Sample images of two RNA from the same family.}
	\label{ref_label_overall}
\end{figure}

\begin{figure}[htp]
    \centering
	\captionsetup[subfigure]{labelformat=empty}
	\subfloat[]{\includegraphics[width=10em]{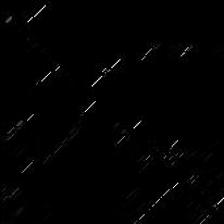}}
	\quad
	\subfloat[]{\includegraphics[width=10em]{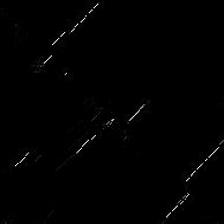}} 
	\caption{Sample images of two RNA from different families.}
	\label{ref_label_overall}
\end{figure}

In Table I, images using all possible combinations of the RNA sequences were generated, except for the training data of different families. With 121 families in the training set, over 2.4 million training images can be generated for sequences from different families if use all possible combinations of RNA are used. However, in these experiments, we found that the models converged quickly without using all the training data. Maintaining an unnecessarily large training set is not only overly complicated but increases training time increasing loading time of images. It also decreases the representativeness of the training data Instead, to reduce the size of the training data, a different approach is taken, The following approach was used to decide the training images for the different family class of the training set. First, we randomly select an RNA from each family. Then, we randomly pick an RNA from every other family and combine them together to generate a training image of different families. We repeat this process 20 times and generate a total of 290,400 different family images of the final training set.

\begin{table}[]
    \centering
    \begin{tabular}{|l|l|l|l|}
    \hline
    Set                    & \# of Families       & Class     & Image Count \\ \hline
    \multirow{2}{*}{Train} & \multirow{2}{*}{121} & Same      & 30,522      \\ \cline{3-4} 
                           &                      & Different & 290,400     \\ \hline
    \multirow{2}{*}{Val}   & \multirow{2}{*}{19}  & Same      & 5,374       \\ \cline{3-4} 
                           &                      & Different & 68,882      \\ \hline
    \multirow{2}{*}{Test}  & \multirow{2}{*}{28}  & Same      & 9,520       \\ \cline{3-4} 
                           &                      & Different & 178,402     \\ \hline
    \end{tabular}
	\caption{Final image count of the dataset before the iterative random image selection algorithm was applied.}
\end{table}

\textbf{Siamese Dataset} As discussed in section 3.1.1, there are two options for designing the dataset to be evaluated by CNNs. Either classify the RNA directly to a family, or classify two RNA as ``Different Family'' or ''Same Family.'' Siamese networks are explicitly designed to make comparisons given two different inputs though. However, given that a Siamese network needs two inputs, a different dataset than the one used for the stand-alone CNNs is needed. In fact, building the dataset for the Siamese Network was simpler. The original dot-plot images used to form the stand-alone CNN dataset before being stitched together were used. Each input to the Siamese network consisted of two dot-plot matrices. During training, the model simply selected two RNA at run-time. Thus, it was unnecessary to apply limits on the different-class of the training dataset as with the stand-alone CNN dataset. The exact set of images seen each iteration during training were different. However, both the standard CNN and Siamese networks were validated and tested on the same set of RNA. The sizes of the validation and test sets for the Siamese networks are exactly half the size of the dataset for the CNN. This is due to the reversibility allowed by the stand-alone CNNs that is unnecessary with the Siamese networks. Each image given to the CNNs represents two RNA, but there is a second combination of both RNA that also represents the two RNA. In order to effectively evaluate how well the CNN is at classifying RNA, both cases must be tested. With the Siamese network, flipping the inputs has no effect on the output of the network. Looking at the Siamese network as a function $S(x,y)$, given two RNA, $A$ and $B$, $S(A,B)$  will always give the same output as $S(B,A)$,
\section{Classification Models}

\textbf{Convolutional Neural Networks} In math, convolution is a mathematical operation on two functions that produces a third function that expresses how the shape of one is modified by the other. In image processing, convolutions can be expressed as discrete filters over the image. Using these filters, create many feature maps that relate nearby pixels. This relation allows for localized feature gathering to be expressed in a global classification when used over an entire image. Convolutional neural networks are built utilizing these convolutions. CNNs contain two parts, a feature extractor and a fully-connected network. The feature extractor is composed of stacked convolutional layers mixed with activation and pooling layers. The stacked convolutional layers, serving as feature extractors, are each followed by an activation layer, typically ReLU activation, to allow for the extraction of nonlinear features \cite{li20202}. The pooling layers, which typically come at the end of the convolutional layers, or at the end of a block of convolutional layers depending on the model design, reduce the spatial resolution of the feature maps, making the model more robust to input distortions \cite{convolutional_review}. The fully-connected network is a densely connected network that interprets the feature representations extracted by the feature network.

In the experiment, two classical CNN-based classification models were implemented: VGGnet \cite{vggnet} and ResNet \cite{resnet}. VGGnet was showed how larger convolution filters could be implemented more efficiently using stacked $3  \times 3$ convolutions. VGGnet follows similar previous CNN constructions. Each block of convolutional layers is followed by max pooling layers to select the most important features to pass to the next block of convolutional features. Each convolutional block implements a higher-order convolution used by many previous models but is implemented as stacked $3 \times 3$ layers for efficiency. VGGnet uses a fully-connected network to evaluate the features. This network has two layers of 4096 neurons each and then a final 1000-neuron with softmax activation to make the final classification. It achieved first in the ImageNet Challenge 2014 localization task and second in the classification task with VGGnet-19 scoring 7.5\% top-5 error rate. We selected the largest VGGnet model, VGGnet-19, a network with 16 convolution layers and 3 fully-connected layers, because it has the best performance of all the VGGnet networks.

\begin{figure}[htp]
    \centering
	\captionsetup[subfigure]{labelformat=empty}
	\subfloat[]{\includegraphics[width=25em]{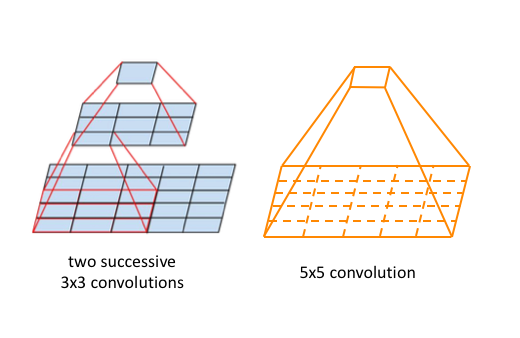}}
	\caption{Convolution Example}
\end{figure}

ResNet \cite{resnet} is an improvement on the VGGnet \cite{vggnet} based scheme. ResNet introduced the concept of residual mapping in the deep learning field. Convolutional Neural networks attempt to approximate a function, but as the network gets deeper it can become increasingly hard to train. By inserting these residual maps throughout the network, ResNet was able to achieve significantly deeper networks than its predecessors. Diverging from many previous models, ResNet models do not use any pooling layers within the convolutional layers. A pooling layer is used at the beginning of the input to reduce the size of the image and highlight the localized features in each pool. The convolutional layers are also followed by a max pooling layer to aggregate the features once they have been extracted. Different from many other CNN, ResNet models use a much more limited fully-connected layer. Unlike VGGnet's two fully-connected layers feeding into a softmax layer to determine the image class, ResNet's convolutional layers output to a 2048-neuron layer that then feeds directly into a softmax layer for classification. As with the VGGnet model, the final softmax layer is changed from a 1000-neuron layer to a 2-neuron layer for the two output classes in this experiment. Two ResNet models were picked, ResNet-50 ResNet-101. ResNet achieved first in the ImageNet 2015 classification task with ResNet-50 and ResNet-101 scoring 5.25\% and 4.60\% top-5 error rate respectively. ResNet also achieved first in the ImageNet 2015 detection and localization tasks. The ResNet and VGGnet architecture can be found in Fig. 8.

\begin{figure}[htp]
    \centering
	\captionsetup[subfigure]{labelformat=empty}
	\subfloat[]{\includegraphics{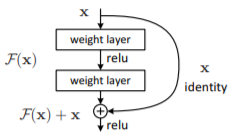}}
	\caption{Residual Block used in ResNet \cite{resnet}}
\end{figure}

\begin{figure}[htp]
    \centering
	\captionsetup[subfigure]{labelformat=empty}
	\includegraphics[]{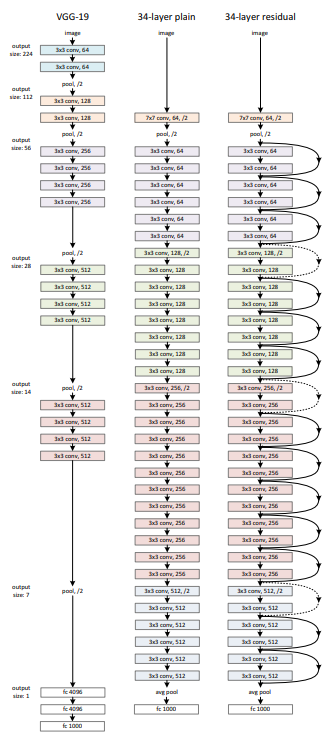}
	\caption{VGGnet-19 and ResNet example \cite{resnet}}
\end{figure}

\subsection{Siamese Networks}
The Siamese network used in these experiments is built on top of the deep CNN models: ResNet-50, ResNet-101, and VGGnet-19 models. These CNNs are run in place with the fully-connected layer removed. Both input images are run through the same CNN but with the fully-connected layers removed. These outputs are then flattened and run through a euclidean distance layer which calculates the absolute value of the difference between both flattened tensors. The output of the euclidean distance layer is then run through a fully-connected network. This framework is more functional for the VGGnet model. The VGGnet model uses a more rigid structure making it much simpler to strip off its fully-connected layers. The ResNet models instead use dimensional reduction throughout the model resulting in a much smaller output size from the convolutional layers. This makes the ResNet model less primed for a Siamese architecture. Simply stripping off the fully-connected layers at the end of the model does not expose the features being gathered by the convolutional layers. The output of the VGGnet feature network is 25,088 features while the output of the ResNet feature network is just 2,048 features. The flattened layers of the two frameworks were also set differently. Because the ResNet was already being flattened as a product of its architecture, the flatten layer of the ResNet models was set to 1,024 neurons. The VGGnet had significantly more features and through experimentation it was found that the size of the flatten layer did not have a noticeable impact on performance, so the VGGnet's Siamese architecture had a flatten layer of just 256 neurons. The Euclidean layer maintained the number of features coming from the flatten layer. The fully-connected network had two layers with the same number of features as output from the euclidean layer (1,024 for ResNet models and 256 for the VGGnet model) followed by a third layer which was a single neuron with Sigmoid activation to determine if the two images were from different families or the same family.

\begin{figure}[htp]
    \centering
	\captionsetup[subfigure]{labelformat=empty}
	\subfloat[]{\includegraphics[width=17em]{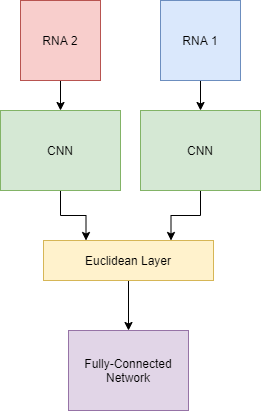}}
	\caption{Siamese Network Architecture}
\end{figure}

\section{Experimental Results}

\textbf{Convolutional Neural Networks} In the first experiment we compared the performance of three models: VGGnet-19, ResNet-50, and ResNet-101. All models employ the SGD optimizer with a momentum factor of 0.9. The initial learning rates of ResNet and VGGnet-19 are set at 0.01 and 0.005, respectively, and the learning rates are reduced by a factor of 0.25 every 50 iterations. The batch size is set at 320. At each iteration, the model is trained on 320 images, selected randomly from the training dataset. Each training iteration took 18 seconds (all training was done on an Nvidia k40 GPU). The models are validated every 50th iteration on the entire validation set. Each validation epoch took 15 minutes. Training for each model took approximately six hours. Both ResNet models were trained for 600 iterations, and VGGnet-19 was trained for 700 iterations. After training, we choose the model with the best performance on the validation set. Then, we evaluate their performance using the average-class accuracy on the test set.

\begin{table}[]
	\begin{tabular}{|l|l|c|c|c|}
		\hline
		\multicolumn{5}{|c|}{Accuracy}                                                                                                                      \\ \hline
		\multicolumn{2}{|l|}{Model}                        & \multicolumn{1}{l|}{VGG-19} & \multicolumn{1}{l|}{ResNet-50} & \multicolumn{1}{l|}{ResNet-101} \\ \hline
		\multicolumn{5}{|c|}{Diff-Same Ratio: 1-1}                                                                                                          \\ \hline
		\multicolumn{2}{|l|}{Train}                        & 92.3\%                        & 94.7\%                           & 96.9\%                            \\ \hline
		\multicolumn{2}{|l|}{Val}                          & 78.2\%                        & 78.8\%                           & 80.0\%                            \\ \hline
		\multicolumn{1}{|c|}{\multirow{3}{*}{Test}} & Diff & 80.0\%                        & 83.0\%                           & 84.0\%                            \\ \cline{2-5} 
		\multicolumn{1}{|c|}{}                      & Same & 85.0\%                        & 82.0\%                           & 83.5\%                            \\ \cline{2-5} 
		\multicolumn{1}{|c|}{}                      & Avg  & 82.5\%                   & 82.5\%              & \textbf{83.5\%}                 \\ \hline
		\multicolumn{5}{|c|}{Diff-Same Ratio: 2-1}                                                                                                          \\ \hline
		\multicolumn{2}{|l|}{Train}                        & 91.9\%                        & 92.8\%                           & 96.6\%                            \\ \hline
		\multicolumn{2}{|l|}{Val}                          & 79.2\%                        & 81.5\%                           & 80.8\%                            \\ \hline
		\multirow{3}{*}{Test}                       & Diff & 84.0\%                        & 86.0\%                           & 86.0\%                            \\ \cline{2-5} 
		& Same & 84.0\%                        & 84.0\%                           & 83.0\%                            \\ \cline{2-5} 
		& Avg  & 84.0\%                      & \textbf{85.0}\%                           & 84.5\%                    \\ \hline
		\multicolumn{5}{|c|}{Diff-Same Ratio: 4-1}                                                                                                          \\ \hline
		\multicolumn{2}{|l|}{Train}                        & 90.0\%                        & 95.6\%                           & 95.6\%                            \\ \hline
		\multicolumn{2}{|l|}{Val}                          & 77.5\%                        & 80.5\%                           & 80.0\%                            \\ \hline
		\multirow{3}{*}{Test}                       & Diff & 87.0\%                        & 89.0\%                           & 90.0\%                            \\ \cline{2-5} 
		& Same & 78.0\%                        & 81.0\%                           & 80.0\%                            \\ \cline{2-5} 
		& Avg  & 82.5\%                        & \textbf{85.0\%}                & \textbf{85.0\%}                 \\ \hline
	\end{tabular}
	\caption{Comparative results of different models trained on different dataset ratios.}
\end{table}

As shown in Table I, the two classes of the dataset are highly imbalanced. The size of different-family images is much larger than that of the same family. In order to reduce the influence of the class imbalance in the training set, different class ratios are chosen for the training stage. Three Different-same ratios were chosen: 1:1, 2:1, and 4:1, as shown in Table II. At the ratio of 1:1, for the batch of 320 training images at each iteration, an equal number of different-class and same-class images are chosen. However, at the ratio of 4:1, 256 and 64 images are selected randomly from the classes of different-family and same-family respectively so as to ensure more different-family images are involved in the training process. The different ratios only apply to the training, while for validation and test, no class ratios are used. Instead, the entire validation and test sets are during their respective phases. The comparative results can be found in Table 4.1.

\textbf{Siamese Network} The second experiment was done with Siamese networks. Both ResNet models, ResNet-50 and ResNet-101, as well as VGGnet-19 were tested within the framework of the Siamese network. Each model was trained for 400 iterations with each training iteration feeding 3200 image pairs to the Siamese network. Every 25 iterations the model was validated on the entire validation set. Each model was trained with an initial learning rate of 0.001 which was decreased to 0.0002 after 300 iterations. The most successful iteration determined by the validation accuracy was then tested on the entire test set. Training the Siamese models took significantly longer, requiring about 20 hours to fully train. As with the CNN models, multiple different-same ratios were also experimented with during training of the Siamese networks. The models were initially trained with a 1:1 ratio, but given the high training accuracy of the VGGnet model, more extreme training ratios were used. As well as the 1:1 ratio, a 9:1 ratio and no ratio were also tested. No controls were applied to the validation or test sets. Full Siamese network results can be found in Table III.

\begin{table}[h]
    \centering
    \begin{tabular}{|c|c|l|l|l|}
        \hline
        \multicolumn{2}{|c|}{Model}       & \multicolumn{1}{c|}{VGG-19} & \multicolumn{1}{c|}{ResNet-50} & \multicolumn{1}{c|}{ResNet-101} \\ \hline
        \multicolumn{5}{|c|}{Different-Same Ratio: 1-1}                                                                                    \\ \hline
        \multicolumn{2}{|c|}{Train}       & 98.9\%                      & 77.3\%                         & 74.3\%                          \\ \hline
        \multicolumn{2}{|c|}{Val}         & 72.5\%                      & 60.1\%                         & 62.8\%                          \\ \hline
        \multirow{3}{*}{Test} & Different & 86.5\%                      & 74.4\%                         & 75.4\%                          \\ \cline{2-5} 
                              & Same      & 70.9\%                      & 73.0\%                         & 78.4\%                          \\ \cline{2-5} 
                              & Average   & 78.7\%                      & 73.7\%                         & 75.4\%                          \\ \hline
        \multicolumn{5}{|c|}{Different-Same Ratio: 9-1}                                                                                    \\ \hline
        \multicolumn{2}{|c|}{Train}       & 96.5\%                      & 73.7\%                         & 71.7\%                          \\ \hline
        \multicolumn{2}{|c|}{Val}         & 72.9\%                      & 61.0\%                         & 61.8\%                          \\ \hline
        \multirow{3}{*}{Test} & Different & 83.5\%                      & 70.5\%                         & 73.4\%                          \\ \cline{2-5} 
                              & Same      & 72.9\&                      & 80.7\%                         & 76.6\%                          \\ \cline{2-5} 
                              & Average   & 78.2\%                      & 75.6\%                         & 75.0\%                          \\ \hline
        \multicolumn{5}{|c|}{Different-Same Ratio: No ratio}                                                                               \\ \hline
        \multicolumn{2}{|c|}{Train}       & 95.7\%                      & 75.6\%                         & 76.0\%                          \\ \hline
        \multicolumn{2}{|c|}{Val}         & 73.0\%                      & 66.5\%                         & 62.1\%                          \\ \hline
        \multirow{3}{*}{Test} & Different & 82.4\%                      & 75.6\%                         & 73.8\%                          \\ \cline{2-5} 
                              & Same      & 76.2\&                      & 72.2\%                         & 76.3\%                          \\ \cline{2-5} 
                              & Average   & 79.3\%                      & 74.9\%                         & 75.1\%                          \\ \hline
    \end{tabular}
	\caption{Comparative results of different Siamese networks trained on different dataset ratios.}
\end{table} 

There is a big difference between the ratio used to control the training dataset of the stand-alone CNN models compared to the Siamese networks. With the stand-alone CNN, the model is learning to evaluate two different RNA within the same feature network. The CNN is being trained to learn features in each half of the image and then using the fully-connected network to evaluate the similarity between these two patches. For this reason, the model needs to be fed a similar number of images from each class to learn the distinct features between the same and different classes. To moderate this learning process, training ratios of 1:1, 2:1, and 4:1 were chosen. When conducted with the 1:1 training ratio, this experiment can evaluate how well the stand-alone CNN approximates a balanced dataset. With the 4:1 training ratio, this experiment evaluates how well the stand-alone CNN approximates a larger, but less balanced dataset. The 2:1 training ratio was also included in the experiment to provide more data on how the balance of the dataset affects the efficacy of the model.

With the Siamese network, the feature network does not need to worry about identifying differences along with the split between the two RNA. Instead, it just needs to learn the important features provided by the dot-plot of the secondary structure. Seeing more unique RNA during training will improve this feature network without fear of the classifier associating each RNA's features with the same family or different family class. For this reason, a more aggressive training ratio was experimented with. The 1:1 ratio was kept to provide control and allow for more direct comparisons to be made between the training process of the stand-alone CNN and the Siamese network. With that, no set training ratio was experimented with to evaluate how well the Siamese network learns the RNA features. The 9:1 ratio was also experimented with to allow for more complex trends to be seen between the 1:1 training ratio and no training ratio.

\textbf{Evaluation Metrics} There are several common methods for evaluating the success of neural networks. It is necessary to choose the right techniques to evaluate the network, based on the goals of the model. Frequently used evaluation metrics are: accuracy, sensitivity, specificity, and F-score. In binary classification, these evaluation metrics are defined by the models' ability to correctly guess positive and negative samples correctly. Correctly labeling a positive is a true-positive (TP), mislabeling a negative as a positive is a false-positive (FP), correctly labeling a negative is a true-negative (TN), and mislabeling a positive as a negative is a false-negative (FN). Using these four definitions, all the evaluation metrics can be defined:

\begin{center}
    $Accuracy = \frac{TP + TN}{FN + FP + TN + TP}$
\end{center}

\begin{center}
    $Sensitivity = \frac{TP}{FN + TP}$
\end{center}

\begin{center}
    $Specificity = \frac{TN}{FP + TN}$
\end{center}

\begin{center}
    $F$-$score = 2*\frac{sensitivity \times specificity}{sensitivity + specificity}$
\end{center}

For these experiments, accuracy, specificity, and sensitivity were used. One other metric, average-class accuracy, was also used. For this thesis, specificity will be defined as the accuracy of the model when given two RNA from different families; sensitivity is the accuracy of the model when given two RNA from the same family; and average-class accuracy is defined for this paper as the average of the different-class and same-class accuracies. Because the test set for these experiments is very unbalanced, using pure accuracy as a metric would be deceptive to the performance of the model. A model could get over 94\% accuracy by always classifying two RNA as different-class. Instead, we use the metric total-accuracy, defined for this paper as the accuracy of the model on the entire test dataset. Average-class accuracy is important because of the significantly larger size of the different-class data versus the same-class data. Were a model to always predict different-class, it could achieve a total accuracy of 95\%. But, for the same reason, a model should predict different-class most of the time because the likelihood of two RNA being from the same family is so incredibly low. For that reason, we also take into account total accuracy and look at the individual class accuracies to evaluate how well these models would work. A false-positive, indicating two RNA were predicted to be from the same class but not, would drastically hinder a clustering algorithm from correctly identifying families compared to a false-negative, indicating two RNA were predicted to be from a different class but weren't. A high total accuracy combined with a low sensitivity indicates that the model is much more likely to predict different-class as opposed to same-class. This is preferred to the inverse as it indicates that the model is right a high percentage of the time when predicting the same-class.

\section{Results}

\begin{figure}[htp]
    \centering
	\captionsetup[subfigure]{labelformat=empty}
	\subfloat[]{\includegraphics[width=8.2em]{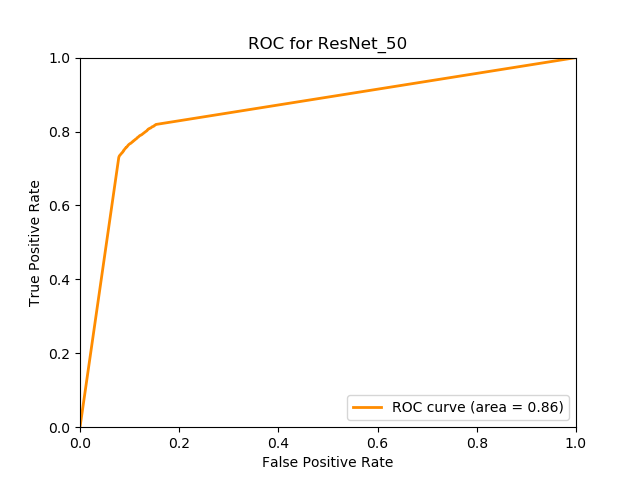}}
	\subfloat[]{\includegraphics[width=8.2em]{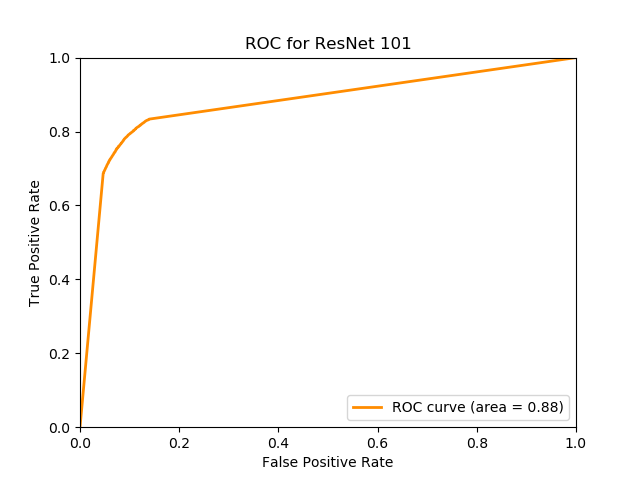}}
	\subfloat[]{\includegraphics[width=8.2em]{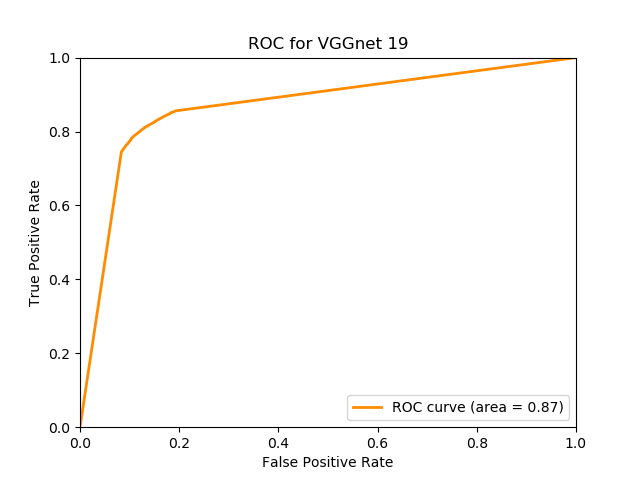}}\\
	\caption{Receiver Operating Curves for CNN models.}
\end{figure}

\begin{figure}[htp]
    \centering
    \captionsetup[subfigure]{labelformat=empty}
    \subfloat[]{\includegraphics[width=8.2em]{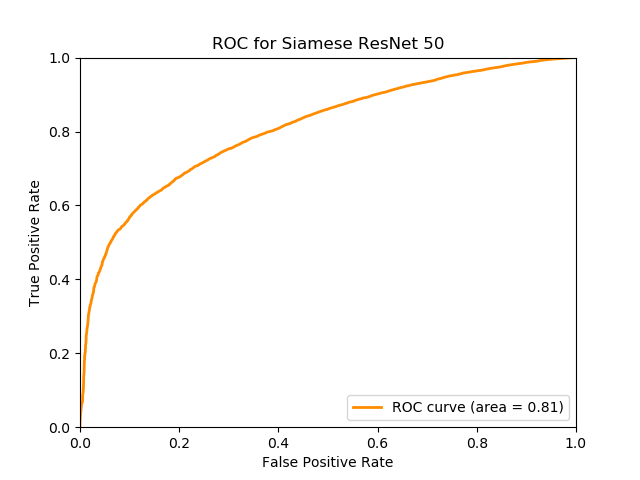}}
	\subfloat[]{\includegraphics[width=8.2em]{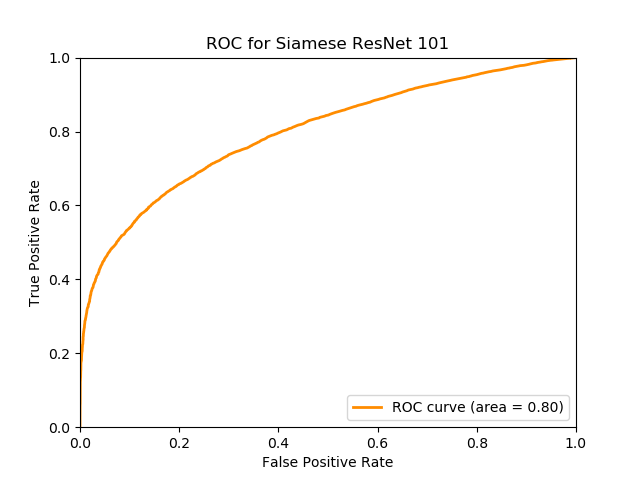}}
	\subfloat[]{\includegraphics[width=8.2em]{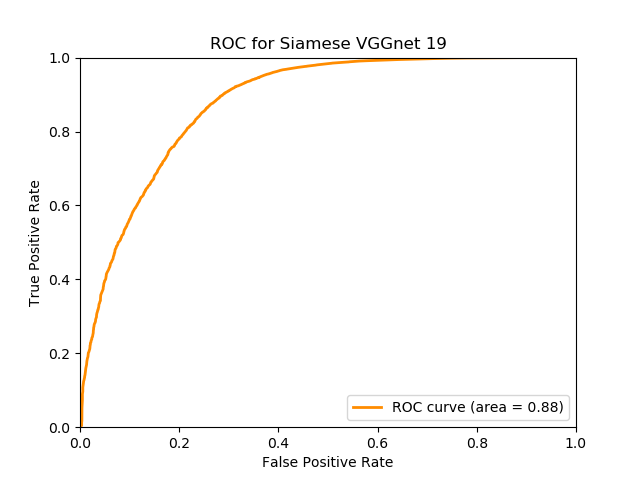}}
	\caption{Receiver Operating Curves for Siamese models.}
\end{figure}

Both ResNet CNN models achieved an average-class accuracy of 85.0\% for ResNet-50 and ResNet-101 respectively. The VGGnet-19 model performed slightly worse with a top accuracy of 82.5\%. The best total accuracy of the CNN models was achieved by ResNet-101 with a total accuracy of 89.5\% on the test set while the ResNet-50 model achieved a slightly worse 88.6\%. As evidenced in the results, a higher Diff:Same ratio resulted in the model improving its accuracy on the different classes while decreasing its performance on the same class. However, even with the decrease in sensitivity, both the average-class accuracy and total accuracy increased for all models. The CNN models saw a performance drop when using the Siamese architecture. All three models had a reduced average-class accuracy when utilizing the Siamese structure, but the VGGnet-19 model was the best performing model when used in the Siamese architecture. The two ResNet models saw a significant reduction in training, validation, and test accuracies while seeing a much smaller reduction in testing accuracy, but the VGGnet model achieved higher training accuracy than with the stand-alone CNN model. There was not a significant change in average-class accuracy across the different training class-ratios for the Siamese networks as there was with the CNN models. The difference between the performance of the VGGnet Siamese model and the ResNet Siamese models is likely due to their structure. For this experiment, only the fully-connected layers were removed for the feature-gathering part of the Siamese structure. However, the ResNet models shrink their output feature size at their base, reducing to only 2048 features. The VGGnet model reduces to a significantly larger 25,088 features. This leaves many more features to be passed to the euclidean layer and gives the VGGnet model much more power in the Siamese structure. With more experimentation and hands-on control over the exact layout of the ResNet models, the ResNet Siamese models should improve based on their better performance as stand-alone networks.

\section{conclusion}

It is difficult to directly compare the output from the neural networks to the output of other methods because most other methods look at classification of very specific RNA families and compare classification accuracy based on pairwise sequence identity. These approaches ignore pairwise sequence identity and just look at strict classification accuracy. The closest approach in the literature is the most recent CNNclust \cite{cnnclust}, which employs one-dimensional convolutions on the primary and secondary sequences to look for common patterns in both. CNNclust only achieved 75.2\% clustering accuracy when tested on ncRNA families not from the training set \cite{cnnclust}. This method is able to achieve better performance by utilizing two-dimensional convolutions on just the secondary structure. Once the models are trained, the models can process around 80 RNA comparisons per second. However, traditional approaches are very time-consuming since they have to perform pairwise alignment on their base-pairing probability matrices, which has a time complexity of $O(l^4)$ while $l$ is the lengths of the ncRNA sequences.

These studies demonstrate the potential of applying CNN-based image classification models to RNA classification. In order to further improve these results, a few measures can be taken. One approach is to increase the dataset to include more RNA families by removing the limitations on sequence length. The distortion caused by increasing the disparity in sequence length may lower the accuracy. However, RNA sequence length amongst families does not vary much so it is more likely that the CNN's feature network recognizes this distortion as a strong indicator of different classes. Another possible measure is applying a clustering algorithm to the classification results. These clusters can be determined by groups of RNA with a high percentage of same-family classifications amongst themselves. Clustering would also give more meaningful data to tune the hyper-parameters. We can also adopt a better approach to handle the imbalance issue of the dataset. As shown in Table II, when we choose 4-1 Diff:Same ratio, the test accuracy for different families is close to 90\%, which means there is still room to further increase the performance of the CNN models. CNNs effectiveness as feature extractors on RNA secondary structure is demonstrated by their 85\% test accuracy in this experiment. With more work, the Siamese networks, which utilize the CNNs for feature extraction, should be able to see at minimum comparable results.

Further tuning of the ResNet models should yield better results within the Siamese architecture. One further study that can be done with little modification is removing the flatten layer from the Siamese network. Typically, Siamese networks use a flatten layer between the feature extractor and the euclidean layer to collapse the features and simplify approximation of the classification function on the set of features. However, in this case it is possible that the flatten layer is acting as a small fully-connected network in between the CNN and the euclidean layer, obfuscating the features from the euclidean layer and resulting in worse accuracy. This is especially the case with the ResNet which already collapses the network and removes most of the fully-connected layer entirely. Siamese networks work by using a feature extractor to analyze two different inputs and then compare their features to determine their similarity, something that should work well with the feature gathering capabilities that have been consistently demonstrated by CNNs. The fact that the ResNet models performed worse in a Siamese architecture is more indicative of poor integration than it is that the entire approach is bad. The VGGnet performed only slightly worse in the Siamese architecture compared to the stand-alone environment. In fact, the VGGnet Siamese model had better training accuracy than the stand-alone model. This indicates one of three possibilities. One is that the Siamese network architecture can be tuned and will yield better validation and test accuracy. The other is that the Siamese network architecture overfits on the training dataset, resulting in worse validation and test accuracy. This would also imply that the Siamese model is learning the features of the training set very well which is what is causing the overfitting. Removing the flatten layer and increasing the number of features for the fully-connected layer to classify on should decrease the overfitting and improve the validation and test accuracy. As a stand alone the model's fully-connected layer is evaluating 25,088 features while the Siamese network's fully-connected layers only have 256 features to evaluate. 

In this work, we presented a new approach for classifying RNA based on their secondary structure through image classification. By treating the BPPM representation of the secondary structure as an image, our approach takes advantage of the high speed and powerful feature extraction capabilities of CNNs. We demonstrated this approach to be a promising way to advance RNA analysis by providing a tool for more accurate and faster RNA classification, achieving 85\% classification accuracy on long noncoding RNA. We also demonstrated this approach with Siamese networks and showed that CNNs are better able to analyze RNA secondary structure when the RNA are combined within the same image. The developed dataset can be taken as a benchmark set for any learning-based research on RNA classification.

\section*{Acknowledgement}
The work was supported in part by Nvidia GPU grant. The following people also contributed to this work: Krushi Patel, Usman Sajid, and Wenchi Ma.

\bibliographystyle{IEEEtranS}      % basic style, author-year citations
\bibliography{refs}   % name your BibTeX data base
% Generated by IEEEtranS.bst, version: 1.12 (2007/01/11)

\end{document}